**Using i-vectors for subject-independent cross-session EEG transfer learning**
Authors: Jonathan Lasko, Jeff Ma, Mike Nicoletti, Jonathan Sussman-Fort, Sooyoung Jeong, William Hartmann

**Title: Using i-vectors for subject-independent cross-session EEG transfer learning**

# ABSTRACT

Cognitive load classification is the task of automatically determining an individual's utilization of working memory resources during performance of a task based on physiologic measures such as electroencephalography (EEG). In this paper, we follow a cross-disciplinary approach, where tools and methodologies from speech processing are used to tackle this problem. The corpus we use was released publicly in 2021 as part of the first passive brain-computer interface competition on cross-session workload estimation. We present our approach which used i-vector-based neural network classifiers to accomplish inter-subject cross-session EEG transfer learning, achieving 18% relative improvement over equivalent subject-dependent models. We also report experiments showing how our subject-independent models perform competitively on held-out subjects and improve with additional subject data, suggesting that subject-dependent training is not required for effective cognitive load determination.

# INTRODUCTION

The ability to determine cognitive load has become more important as the complexity of systems increases, and the amount of data available to the system operator increases in kind. Operator performance and the ability to handle multiple tasks in a dynamically changing environment is dependent on levels of applied attention and working memory. In terms of brain processing, this corresponds to so-called top-down processing or increased population-level neuronal firing in higher cortical areas that modulate focus and attention. Modern human-machine interfaces (HMIs) are a critical tool in optimizing operator performance and must strike a balance between providing enough information to allow for optimal situational awareness, while not overwhelming the operator. Objective measurement of the cognitive workload, stress, and health is necessary to understand operator performance while using increasingly complex systems.

Cognitive load refers to the use and management of working memory resources [1]. Conditions in which a human is over-saturated with information can lead to a state of cognitive overload and deterioration of performance in decision-making. Physiologic measures provide a window into the mental and physical state of the user over time. Heart rate, breathing rate, galvanic skin response, eye-tracking, and electroencephalography (EEG) measurements have all been used to measure an individual's cognitive load while performing a task [2]–[6].[2][3][4][5][6] The characterization of these neurophysiologic changes that correspond to working memory, attention, and executive function has been demonstrated in multiple studies, with earlier work focusing on identifying specific time and frequency domain features [7]–[9].[7][8][9] When accurately determined, this measure of load can serve as feedback into HMIs to triage the presentation of information [9]–[11].[9][10][11] This allows for dynamic presentation of information and requests for operator input so as not to overload the user during dynamic periods of increased stress, while maintaining vigilance during longer static periods.

**Using i-vectors for subject-independent cross-session EEG transfer learning**
Authors: Jonathan Lasko, Jeff Ma, Mike Nicoletti, Jonathan Sussman-Fort, Sooyoung Jeong, William Hartmann

Approaches to cognitive load classification rely on both onset prediction and model individualization. Given this fact, previous work has demonstrated the difficulties in cross-session prediction due to the variability and non-stationarity of neurophysiologic signals and changes in experimental conditions or underlying neurophysiologic states across days or even hours [12]–[14].[12][13][14]Typical ML/AI approaches have also required EEG sensor training data from the individual under test, drastically limiting the usability of the system for larger populations or as a tool for non-research applications. We demonstrate classification of cognitive workload, specifically cross-session cognitive load prediction, using an approach that also generalizes across subjects.

In this paper we build on BBN's successes applying our mature speech and language processing techniques [15] in other domains [16] by applying these techniques on open-source, time-series EEG workload data. BBN first reported on using i-vectors for EEG classification of emotional responses in 2014 [17]. Subsequent use of i-vectors for EEG classification has focused on subject verification [18]–[20].[18][19][20] In this paper, BBN shows that the i-vector technique can achieve subject-independent cross-session transfer learning for cognitive load prediction.

# METHODS

## Participants and Experimental Paradigm

We obtained an open-source dataset from the Passive Brain-Computer Interface Competition (Hackathon) on intersession workload estimation for the 3$^{rd}$ International Neuroergonomics conference [14]. We chose this dataset because at the time it was the most comprehensive, publicly available set of EEG recordings obtained while eliciting multiple levels of cognitive workload in a user.[1]

The Hackathon dataset consists of pre-processed EEG recordings from 15 participants (6 females; average age 25 years old) who attended three experimental sessions each separated by 1 week. The dataset includes experimental sessions chosen to elicit specific levels of cognitive workload for an operator. Each experimental session consisted of training and a 1-minute resting state measurement with eyes open, followed by a three 5-minute blocks (pseudo-randomized in order by difficulty) in which participants completed a MATLAB version of the NASA MATB-II (Multi-Attribute Battery) task (Figure 1) [21], [22].

---

[1] Subsequently, COG-BCI, a superset of the hackathon dataset, was released by the same team [27]. Our preliminary experiments with the MATB-II data for subjects 16-29 of COG-BCI indicated problems in the data; we later learned these problems had been resolved in the Version 4 release (December 2022) [28].

**Using i-vectors for subject-independent cross-session EEG transfer learning**
Authors: Jonathan Lasko, Jeff Ma, Mike Nicoletti, Jonathan Sussman-Fort, Sooyoung Jeong, William Hartmann

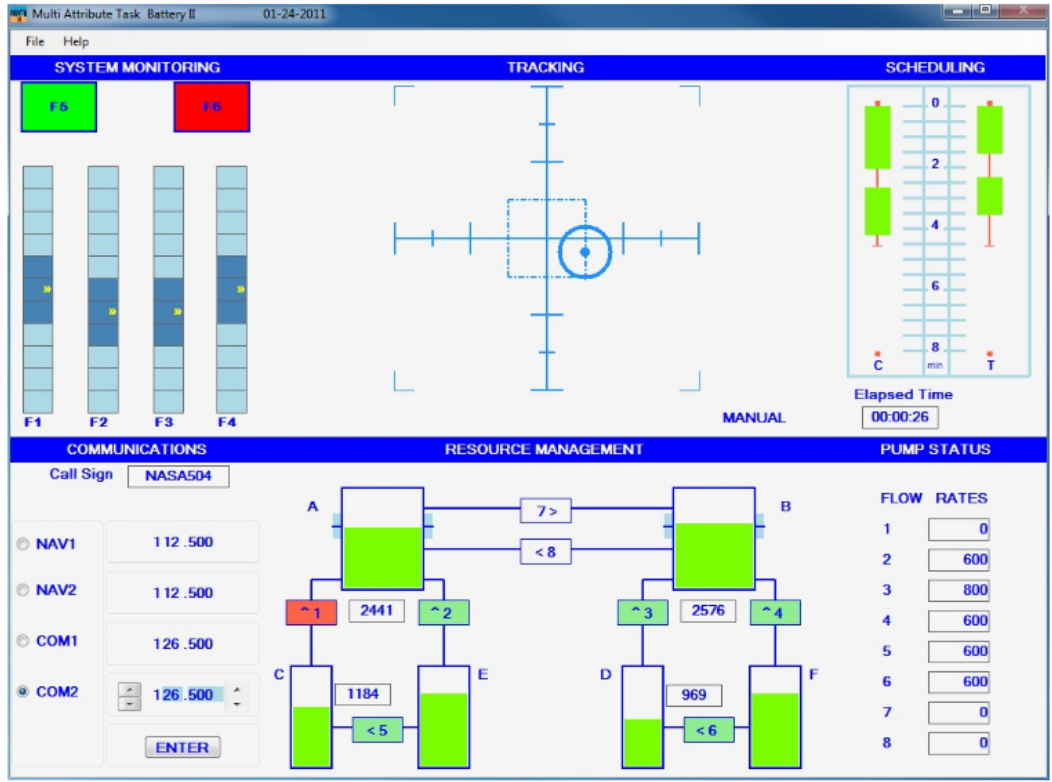

*Figure 1: MATB-II example MATLAB version showing multiple subtasks to be completed by user [16].*

In MATB-II, participants complete visual, auditory, motor, and planning subtasks, which have shown by both objective and subjective measures to elicit different levels of workload for a user. The workload increases with the increasing number of subtasks and underlying difficulty of the tasks required for the user to complete (Table 1) [14], [22].

*Table 1: MATB-II Difficulty levels with corresponding sub-tasks completed by the user.*

| SUB-TASK | Tracking and System Monitoring | Resource Management | Communication |
|---|---|---|---|
| **TASK LEVEL** | | | |
| **Easy** | x | | |
| **Medium** | x | x | |
| **Difficult** | x* | x | x |

*Note that for the Difficult level the tracking task also had an increased target motion speed

## ML Classification Approaches

BBN experimented with two different approaches to fatigue level classification: residual networks (ResNets) trained on raw sensor data and neural networks trained on i-vector features.

### ResNets

**Using i-vectors for subject-independent cross-session EEG transfer learning**
Authors: Jonathan Lasko, Jeff Ma, Mike Nicoletti, Jonathan Sussman-Fort, Sooyoung Jeong, William Hartmann

BBN trained subject-specific ResNet-18 models using the raw data from all 61 channels, using Session 1 data as training and Session 2 as validation. Our ResNet implementation was in PyTorch. We selected an 18-layer deep model due to the limited amount of data. Batch size was 500. Table 2 shows the best classification accuracies achieved on the validation set across 200 training epochs. These low scores indicate that the ResNets were unable to learn features discriminatively from raw data. We did not submit results from these models for Session 3 assessment.

*Table 2: Classification accuracies of ResNet subject-specific models, selected based on best performance on the validation set.*

| Subject | Training (Session 1) | Validation (Session 2) |
|---|---|---|
| All (avg) | 38.66 | 37.88 |
| P01 | 38.26 | 40.05 |
| P02 | 43.18 | 47.20 |
| P03 | 41.61 | 41.61 |
| P04 | 32.89 | 36.47 |
| P05 | 45.41 | 41.61 |
| P06 | 50.78 | 46.09 |
| P07 | 39.60 | 33.78 |
| P08 | 33.56 | 33.56 |
| P09 | 35.35 | 34.89 |
| P10 | 42.28 | 38.48 |
| P11 | 33.56 | 34.00 |
| P12 | 38.93 | 36.47 |
| P13 | 33.33 | 35.12 |
| P14 | 33.33 | 33.33 |
| P15 | 37.81 | 35.57 |

I-Vector Neural Networks

BBN's subject-independent neural network models use a speaker and language recognition technique called i-vectors which are "an elegant way of reducing the large-dimensional variable-length input data to a small-fixed-dimensional feature vector while retaining most of the relevant information" [23]. A more detailed explanation of the i-vector technique is available in [16].

To train these models, we utilized data from all 15 subjects, treating each 2-second epoch as a training sample consisting of 500 frames. Each frame is treated as a 30-dimension feature vector, where the elements of that vector are either (a) the raw values from a 31-channel subset of the 61 channels or (b) the results max pooling or average pooling based on 21- or 25-sub-region groupings of channels, grouped (mostly) by proximity. The exact sensor subset and groupings are shown in Table 3.

**Using i-vectors for subject-independent cross-session EEG transfer learning**
Authors: Jonathan Lasko, Jeff Ma, Mike Nicoletti, Jonathan Sussman-Fort, Sooyoung Jeong, William Hartmann

Next, we trained i-vector extractors for 80-dimensional i-vectors using the aforementioned 30-dimensional feature vectors plus their first-order deltas as input features. The i-vector universal background model had 512 components. We then used these i-vectors as inputs to train neural network classifiers for the three MATB-II levels (easy, medium, and difficult). To smooth out noise in the signal, we utilized smoothing with a moving average (SMA) with a window of either 16 or 20 frames (denoted below as SMA16 and SMA20, respectively) followed by global mean and variance normalization (GMVN).

*Table 3: Our 31-channel subset and the other subregions we used in our max or average pooling approaches.*

| 31-channel subset | 21 sub-regions | 25 sub-regions |
|---|---|---|
| FP1, FP2 | FP1, FP2 | FP1, FP2 |
| Fz | Fz, F1, F2 | Fz |
| F3, F7 | F3, F5, F7 | F1, F3, F5, F7 |
| F4, F8 | F4, F6, F8 | F2, F4, F6, F8 |
| FCz, FC1, FC2 | FCz, FC1, FC2 | FCz |
| FC5 | FC3, FC5 | FC1, FC3, FC5 |
| FC6 | FC4, FC6 | FC2, FC4, FC6 |
| C3 | C1, C3, C5 | C1, C3, C5 |
| C4 | C2, C4, C6 | C2, C4, C6 |
| CP1, CP5 | CP1, CP3, CP5 | CP1, CP3, CP5 |
| CP2, CP6 | CP2, CP4, CP6 | CP2, CP4, CP6 |
|  | CPz | CPz |
| Pz | Pz, P1, P2 | Pz |
| P3, P7 | P3, P5, P7 | P1, P3, P5, P7 |
| P4, P8 | P4, P6, P8 | P2, P4, P6, P8 |
| Oz, O1, O2 | Oz, O1, O2, O3, O4, O5 | Oz, O1, O2, O3, O4, O5 |
| AF7 | AFz, AF3, AF7 | AFz |
|  |  | AF3, AF7 |
|  | AF4, AF8 | AF4, AF8 |
|  | POz, PO3, PO4, PO7, PO8 | POz |
|  |  | PO3, PO7 |
|  |  | PO4, PO8 |
|  |  | T7 |
| T7, FT9 | FT7, T7, TP7 | FT7, FT8 |
| FT8, T8 | FT8, T8, P8 | TP7, TP8 |

While developing the models, we used the Session 1 data as training samples and measured classification accuracies on the Session 2 data. For our final submitted models, we used the combined Session 1 and Session 2 data as the training set. Following this approach, we trained seven subject-independent i-vector neural network classifier systems. The best of these systems (denoted maxP21-SMA16 below) used max pooling over 21 sub-regions of sensors and SMA16. We also combined outputs from these systems (via simple voting method) into a 7-system combination which achieved our best performance on the task overall.

**Using i-vectors for subject-independent cross-session EEG transfer learning**
Authors: Jonathan Lasko, Jeff Ma, Mike Nicoletti, Jonathan Sussman-Fort, Sooyoung Jeong, William Hartmann

*Table 4: Models that made up our final 7-system combination*

| Model name | Sensors (see Table 3) | Pooling | Smoothing |
|---|---|---|---|
| sd31-SMA16 | 31-channel subset | N/A | SMA16 |
| avgP25-SMA16 | 25 sub-regions | average | SMA16 |
| maxP25-SMA16 | 25 sub-regions | max | SMA16 |
| avgP21-SMA16 | 21 sub-regions | average | SMA16 |
| maxP21-SMA16 | 21 sub-regions | max | SMA16 |
| avgP21-SMA20 | 21 sub-regions | average | SMA20 |
| maxP21-SMA20 | 21 sub-regions | max | SMA20 |

As a baseline for comparison with our subject-independent systems, we also trained subject-dependent neural network classifiers for each of the 15 subjects using the same architecture and approach as with our best single system.

# Results

**I-vector neural networks achieve inter-subject transfer learning**

BBN's i-vector neural network classifiers achieved inter-subject transfer learning on this dataset. Comparing the performance of our subject-dependent and single-best models when trained on Session 1 data and validated on Session 2 data, we found that our best single subject-independent model (maxP21-SMA16) outperforms subject-dependent models for the 11 of the 15 subjects (P05, P08, P09, and P10 being exceptions). Pooling on sensors divided into 21 sub-regions outperformed pooling of 25 sub-regions, and max pooling performed slightly better than average pooling on the 21 sub-regions.

Table 5 shows the performance of three representative systems when trained on data from Session 1 and Session 2 and tested on the held-out Session 3 data. As expected, the single-best system incorporating data from all 15 subjects outperformed the subject-dependent systems, and the 7-system combination of subject-independent sub-systems achieved our best score overall. Our system ranked second after the highest score reported from the 2021 Neuroergonomics [4] [12] Passive BCI hackathon (54.26%) [14] and third when considering all scores published on this dataset to date (c.f. 54.3% from [24]).

**Using i-vectors for subject-independent cross-session EEG transfer learning**
Authors: Jonathan Lasko, Jeff Ma, Mike Nicoletti, Jonathan Sussman-Fort, Sooyoung Jeong, William Hartmann

*Table 5: Classification accuracies of i-vector neural network models on Session 3 data*

| Subject | subject-dependent | single-best (maxP21-SMA16) | 7-system combination |
|---|---|---|---|
| **Overall** | **0.44** | **0.48** | **0.52** |
| P01 | 0.41 | 0.53 | 0.55 |
| P02 | 0.57 | 0.64 | 0.68 |
| P03 | 0.44 | 0.54 | 0.58 |
| P04 | 0.52 | 0.51 | 0.56 |
| P05 | 0.43 | 0.50 | 0.51 |
| P06 | 0.39 | 0.41 | 0.46 |
| P07 | 0.43 | 0.50 | 0.54 |
| P08 | 0.42 | 0.46 | 0.46 |
| P09 | 0.46 | 0.41 | 0.53 |
| P10 | 0.32 | 0.50 | 0.56 |
| P11 | 0.45 | 0.40 | 0.44 |
| P12 | 0.52 | 0.51 | 0.56 |
| P13 | 0.46 | 0.41 | 0.48 |
| P14 | 0.46 | 0.43 | 0.49 |
| P15 | 0.39 | 0.43 | 0.46 |

**Performance of subject-independent models on held-out subjects**

BBN also investigated how our subject-independent models performed on users held-out from the original training set. To do this, we trained a subject-independent model using 10 subjects from the training set and tested its performance on the 5 held-out subjects. We did this for two different divisions of train/test. (See Table 6 and Table 7.) These i-vector neural network classifiers were all trained using the 31-channel subset listed in the first column Table 3. No pooling was used in this experiment.

*Table 6: Accuracy of subject-independent models for subjects not seen in training*

| Subject | **Subject-dep Session 2** | Train P04-P13 (Session 1 only) | | Train P04-P13 (Session 1 + 2) | |
|---|---|---|---|---|---|
| | | Session 1 | **Session 2** | Session 1 | **Session 2** |
| **Overall** | **0.38** | **0.42** | **0.46** | **0.45** | **0.44** |
| P01 | 0.36 | 0.46 | 0.47 | 0.48 | 0.43 |
| P02 | 0.41 | 0.38 | 0.47 | 0.44 | 0.37 |
| P03 | 0.42 | 0.44 | 0.46 | 0.44 | 0.53 |
| P14 | 0.36 | 0.42 | 0.41 | 0.43 | 0.36 |
| P15 | 0.35 | 0.43 | 0.48 | 0.48 | 0.52 |

**Using i-vectors for subject-independent cross-session EEG transfer learning**
Authors: Jonathan Lasko, Jeff Ma, Mike Nicoletti, Jonathan Sussman-Fort, Sooyoung Jeong, William Hartmann

*Table 7: Accuracy of subject-independent models for subjects not seen in training. (Different subject selection from previous table.)*

| Subject | **subject-dep Session 2** | Train P1-P10 (Session 1 only) | | Train P1-P10 (Session 1 + 2) | |
|---|---|---|---|---|---|
| | | Session 1 | **Session 2** | Session 1 | Session 2 |
| **Overall** | **0.34** | **0.40** | **0.45** | **0.41** | **0.43** |
| P11 | 0.33 | 0.42 | 0.48 | 0.43 | 0.36 |
| P12 | 0.43 | 0.47 | 0.41 | 0.47 | 0.49 |
| P13 | 0.25 | 0.31 | 0.40 | 0.30 | 0.38 |
| P14 | 0.36 | 0.41 | 0.40 | 0.40 | 0.41 |
| P15 | 0.35 | 0.38 | 0.45 | 0.43 | 0.52 |

Nine of out ten times, the subject-independent model, trained on Session 1 and tested on Session 2, outperformed the subject-dependent models, even though the former had never seen training data from the subjects under test. (Only the subject-dependent model for subject P12 out-performed the subject-independent model.) Our results also suggest that augmenting the subject-independent model's training dataset with additional sessions from previously-seen subjects is not likely to improve its subject-independency (as the models trained on Session 1 + 2 performed slightly worse on average on Session 2 than the subject-independent models trained on Session 1 only).

## Conclusions

The most important implications of our cognitive load classification results are the use of cross-subject information to predict load across sessions and the ability to make predictions on subjects that were not in the training pool. This stood in contrast to the approaches of the other participants of the Hackathon who did not rely on cross-subject or cross-session transfer learning [14].

One possible reason our i-vector neural network models out-perform our ResNet-18 models may be that there is too little data for the ResNet to learn features from the raw data. The neural networks relying on the i-vectors are very small, meaning that the i-vectors themselves are the primary differentiators of cognitive load. Future work might leverage additional techniques from low-resource speech recognition (such as data augmentation and semi-supervised learning).

In terms of objectively measuring the cognitive state of an operator, our results codify our belief that the approach taken in this paper can be used in technologies where the eventual operator will not be a part of the training set. This implies that product development can be freed from certain privacy concerns and helps make the software more deterministic than if its parameterization is highly dependent on field measurements. It also lessens the need for validation between training and performance. Both of these are important steps to bridging the gap towards fielding actual systems outside of controlled experimental settings.

We also note two caveats regarding the development of a real-time cognitive load measurement system. Firstly, the MATB-II datasets utilize a controlled, experimental load elicitation

**Using i-vectors for subject-independent cross-session EEG transfer learning**
Authors: Jonathan Lasko, Jeff Ma, Mike Nicoletti, Jonathan Sussman-Fort, Sooyoung Jeong, William Hartmann

approach, yet for real-time applications, it would be important to distinguish between measures of vigilance and alertness versus attention and working memory. This addresses the so-called lumberjack effect in which reliance on consistently performing high-level automation has led operators to become less engaged and unable to appropriately manage system collapses, ultimately leading to catastrophic failure [25]. Secondly, the assessment of an operator's state should be context-based. In other words, in certain CONOPS or situations a higher degree of load may be appropriate or even required for success on a given task and not necessitate an intervention.

Our cross-session results suggest a common mechanism for the EEG signals indicating cognitive load between test subjects. Work suggesting a common model for brain networks is prevalent [26]; however there is no gold standard definition of how correlated the cognitive load signal should be between subjects. Other more precise techniques of brain imaging could be used to study this phenomenon to better understand the limitations of cross-session techniques and to inform the design of ambulatory EEG measurement devices to make them more palatable for the user.

## Author Contributions

Jonathan Lasko, Mike Nicoletti, and Jon Sussman-Fort spearheaded the internal research effort which led to this work being accomplished. Jeff Ma performed the i-vector neural network experiments. Jonathan Lasko performed the ResNet experiments with help from Sooyoung Jeong and co-wrote the paper. William Hartmann advised on ResNet and i-vector experiments. Jonathan Sussman-Fort advised on interpretation of EEG signal data and co-wrote the paper.

**Using i-vectors for subject-independent cross-session EEG transfer learning**
Authors: Jonathan Lasko, Jeff Ma, Mike Nicoletti, Jonathan Sussman-Fort, Sooyoung Jeong, William Hartmann

**Using i-vectors for subject-independent cross-session EEG transfer learning**
Authors: Jonathan Lasko, Jeff Ma, Mike Nicoletti, Jonathan Sussman-Fort, Sooyoung Jeong, William Hartmann